\documentclass[10pt,twocolumn,letterpaper]{article}

\usepackage{cvpr}  
\usepackage{amsthm} 

\newtheorem{lemma}{Lemma}
\newtheorem{proposition}{Proposition}
\newtheorem{remark}{Remark}
\usepackage{algorithm}
\usepackage{algorithmic}

\definecolor{cvprblue}{rgb}{0.21,0.49,0.74}
\usepackage[pagebackref,breaklinks,colorlinks,allcolors=cvprblue]{hyperref}

\def\paperID{27CY040001} 
\def\confName{ICCV}
\def\confYear{2027}

\title{Object Detection Based on Distributed Convolutional Neural Networks}

\author{Liang Sun\\
Shandong University of Science and Technology\\
Jinan, 250031 P.R. China\\
{\tt\small liangsun@sdust.edu.cn}}

\begin{document}
\maketitle
\begin{abstract}
Based on the Distributed Convolutional Neural Network(DisCNN), a straightforward object detection method is proposed. The modules of the output vector of a DisCNN with respect to a specific positive class are positively monotonic with the presence probabilities of the positive features. So, by identifying all high-scoring patches across all possible scales, the positive object can be detected by overlapping them to form a bounding box. The essential idea is that the object is detected by detecting its features on multiple scales, ranging from specific sub-features to abstract features composed of these sub-features. Training DisCNN requires only object-centered image data with positive and negative class labels. The detection process for multiple positive classes can be conducted in parallel to significantly accelerate it, and also faster for single-object detection because of its lightweight model architecture. 

\textbf{Index Terms}—Object detection, feature detection, distributed neural networks, convolutional neural networks, feature disentanglement.
\end{abstract}

    
\section{Introduction}
\label{sec:intro}

Object detection is always based on object recognition. One of the most popular object recognition methods is the convolutional neural network (CNN) trained with the cross-entropy loss via backpropagation \cite{lecun1999object,lecun2002gradient,lecun2010convolutional,lecun2015deep}. It has led to a range of object detection methods, such as R-CNN \cite{girshick2014RCNN}, Fast R-CNN \cite{girshick2015fast}, Faster R-CNN \cite{ren2015faster}, Mask R-CNN \cite{he2017mask}, and YOLO \cite{redmon2016you}, all of which are end-to-end models that produce a single output vector representing both the object and its location. Therefore, object features and location are entangled, making object detection challenging. Recently, a new type of CNN, called distributed CNN (DisCNN), has been proposed that extracts only the features of a specific positive class while neglecting those of all other classes \cite{sun2026distributed}. The secret recipe is to use a novel loss function, Negatives to Origin (N2O), which introduces an additional constraint that forces the negatives to Origin, in addition to the cross-entropy loss, so that DisCNN produces zero responses to negative features. Therefore, positive features are disentangled not only from features of other classes but also from location information. This may provide an easy but convincing way to detect objects.

In this work, a novel object detection method based on DisCNN is proposed. Firstly, based on the N2O loss, a DisCNN is trained on a dataset comprising a single positive class and two or more negative classes using backpropagation. During object detection, the large image containing positives and complex backgrounds is first partitioned into small patches using multi-scale sliding windows. The patches are then fed to the DisCNN, after being resized to the training data size. Finally, the modules of the DisCNN output vectors are used as a metric to assess the probability that a patch contains a positive sample or positive features, and by overlapping the best-scoring patches of all multi-scale sliding windows, the positive can be localized. The advantage of the method is that: 1. No location-related labeled data, such as bonding boxes, is required for training, and only object-centered image data, along with positive and negative class labels, suffice. 2. DisCNN is much lighter-weight than a general multi-class CNN, so the positive object detection process can be much faster, and DisCNNs of different positive classes can be deployed on multiple GPUs in a distributed manner to detect the positives in a parallel way to keep it still fast for a multi-positive detection scenario. 3. The detection of the whole object is achieved by the detection of its features on multiple scales; therefore, the recognition and localization of the object can be confidently accurate. 4. Sub-features of the object can also be detected.

\section{Model and training}
\label{sec:formatting}
Before object detection, a DisCNN is first trained based on the N2O loss using backpropagation. The entire training process is consistent with that in the literature \cite{sun2026distributed}, and this chapter primarily introduces the essential technical details.

\subsection{DisCNN architecture}

DisCNN adopts the architecture of VGG model \cite{simonyan2014VGG}, as shown in \cref{tab:Table 1}, except for the absence of a softmax layer, and reduces the number of feature maps to 8, since 8 features may be sufficient for cars, such as wheels, license plates, lights, etc. Therefore a DisCNN for one positive class is lightweight, which reduces the time of both training and detection.

\begin{table}[tb]
  \caption{Model Architecture. The convolutional layer parameters are denoted as “$conv<receptive field size>-<number of channels>$”. 
  }
  \label{tab:Table 1}
  \centering
  \tabcolsep=0.5cm
  \begin{tabular}{@{}lll@{}}
    \toprule
    \hspace{2em} DisCNN \\
    \toprule
    \hspace{2em} input(96x96RGB)\\
    \midrule
    \hspace{2em}Conv3-64\\
    \hspace{2em}Batch Normalization \\
    \hspace{2em}ReLU \\
    \hspace{2em}Max pooling \\
    \midrule
    \hspace{2em}Conv3-32\\
    \hspace{2em}Batch Normalization \\
    \hspace{2em}ReLU \\
    \hspace{2em}Max pooling \\
    \midrule
    \hspace{2em}Conv3-16\\
    \hspace{2em}Batch Normalization \\
    \hspace{2em}ReLU \\
    \hspace{2em}Max pooling \\
    \midrule
    \hspace{2em}Conv3-8 \\
    \hspace{2em}Batch Normalization \\
    \hspace{2em}ReLU \\
    \hspace{2em}Max pooling \\
    \midrule
    \hspace{2em}FC-288\\
    \midrule
    \hspace{2em}FC-128 \\
    \midrule
    \hspace{2em}FC-16 \\    
  \bottomrule
  \end{tabular}
\end{table}

\subsection{Training with N2O loss}
The N2O loss essentially constrains the negatives to the origin, while the positives are mapped to a compact set in a higher-dimensional space, just like the cross-entropy loss. There is a fundamental data requirement for H2O loss as presented in \cref{lem:Lemma 1}.

\begin{lemma}
\label{lem:Lemma 1}
The H2O loss requires that the positive class have no features in common with the negative class in the training data.
\end{lemma}

Meanwhile, for most object detection scenarios, small positives embedded in large backgrounds are more common. Thus, the data produced by uniform patch partitioning may be extremely imbalanced and require more negatives in the training dataset, as described in \cref{lem:Lemma 2}. 

\begin{lemma}
\label{lem:Lemma 2}
For object detection, a considerable number of negatives are required for training data.
\end{lemma}

For example, in accordance with \cref{lem:Lemma 1} and \cref{lem:Lemma 2}, the subset \{car, bird, cat, deer, dog, horse, monkey\} of the STL-10 dataset \cite{coates2011STL,deng2009imagenet} is qualified to train a DisCNN, where the car is regarded as the positive class and the rest are negatives that have no features in common with the car. The trained DisCNN has an important property, as described in \cref{lem:Lemma 3}, which is essential for object detection. 

\begin{lemma}
\label{lem:Lemma 3}
The trained DisCNN detects only positive samples, while negatives receive a zero response.
\end{lemma}

Therefore, based on \cref{lem:Lemma 3}, if a small patch from a large detection scenario can activate DisCNN that produces a non-zero vector, it indicates a positive lies in it. This is where the principal idea of this work comes from.

\section{Object detection}
The DisCNN trained with the N2O loss is now being used for object detection. First, an algorithm is proposed to detect positives
in a multi-positive scenario. Then, by applying it to a one-car detection scenario, we conclude that the output vector module of DisCNN and the presence probabilities of positive-class features are positively monotonic. Finally, the algorithm is evaluated in a real-world traffic scenario to detect multiple cars.

\subsection{Object detection algorithm}

The algorithm is presented as show in \cref{alg:algorithm 1}.

\begin{algorithm}
    \caption{Object detection algorithm}
    \label{alg:algorithm 1}
	\begin{algorithmic}
		\STATE \textbf{Step 1:} Based on the N2O loss, a DisCNN as shown in \cref{tab:Table 1} is initialized and trained using backpropagation on a dataset qualified according to Lemma \ref{lem:Lemma 1} and Lemma \ref{lem:Lemma 2}, with image size of $n \times n$. Assume the large image to be detected on is $l \times m$
        \STATE \textbf{Step 2:} Set the initial sliding window size $sws = min(l,m)$, and minimum sliding window size  $min\_sws$ is roughly chosen by the minimum feature size.
		\STATE \textbf{Step 3:} Set $stride = sws/3$, window attenuation $wa = sws/20$, and partition the large image to be detected into patches with current $sws$ and $stride$.
        \STATE \textbf{Step 4:} Feed the patches into the DisCNN,  which is resized into $n \times n$, and then compute the modules of the output vectors.
        \STATE \textbf{Step 5:} Choose an appropriate threshold $thr$, if $module > thr$, the corresponding patch is positive, which contains positive features.
        \STATE \textbf{Step 6:} Preserve all the location information $xmin$, $ymin$, $xmax$, $ymax$ of the positive patches, and let $sws = sws - wa$. 
        \STATE \textbf{Step 7:} If $sws > min\_sws$, then goto setp3; else goto step 8.          
        \STATE \textbf{Step 8:} Compute the center coordinates of the positive patches, cluster them, and then draw a bonding box for each cluster with [$min(xmins)$, $min(ymins)$, $max(xmaxs)$, $max(ymaxs)$] of each cluster. Thus, the positives in the large image are detected and located. 
	\end{algorithmic}
\end{algorithm}

\begin{remark}
\label{rem:remark 1} 
The training process in \cref{alg:algorithm 1} requires only object-centered image data along with positive and negative class labels; no location-related labels, such as bounding boxes, are required.
\end{remark}

\begin{remark}
\label{rem:remark 1} 
Obviously, \cref{alg:algorithm 1} can be executed repeatedly on datasets of different positive classes, thus enabling multi-class object detection. Then, DisCNNs for different positive classes can be deployed in a distributed manner across multiple GPUs to localize positives in parallel, thereby significantly accelerating detection even when multiple positives of different classes are present. 
\end{remark}

\subsection{Module is a metric for the presence probability of positive features}
\label{3.2}

\begin{figure*}
  \centering
  \begin{subfigure}{0.46\linewidth}
    \includegraphics[height=6.9cm]{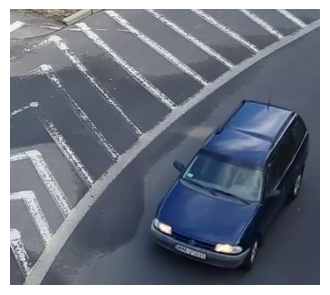}
    \caption{One-car scenario}
    \label{fig:short-a0}
  \end{subfigure}
  \hspace{0mm}
  \begin{subfigure}{0.4\linewidth}
    \includegraphics[height=2cm]{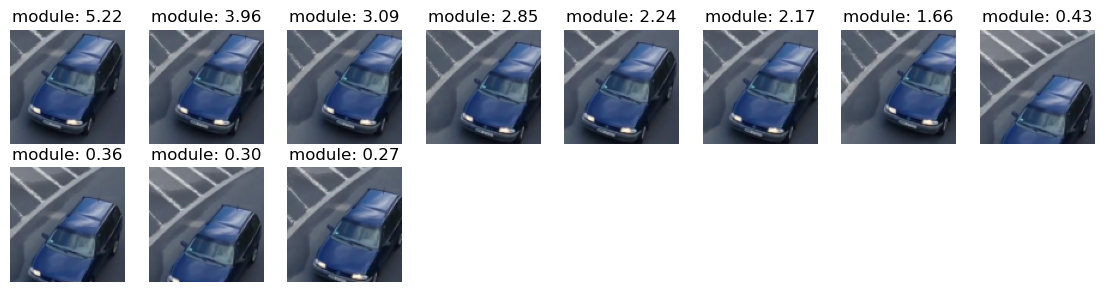}
    \caption{Sliding window size = 280 }
    \label{fig:short-b0}
    
    \includegraphics[height=2cm]{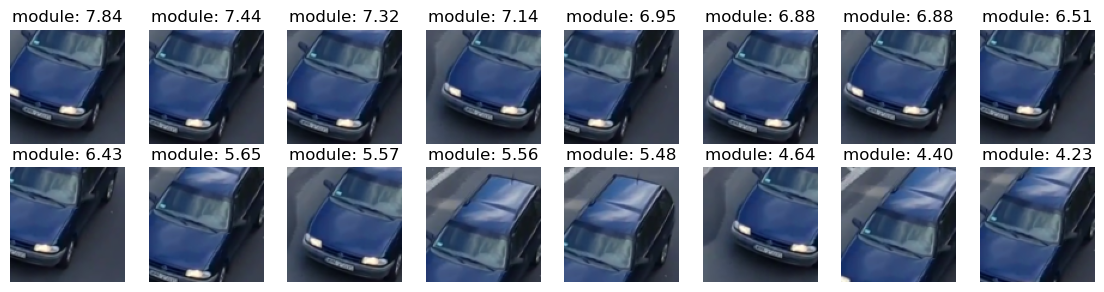}
    \caption{Sliding window size = 180}
    \label{fig:short-c0}
    
    \includegraphics[height=2cm]{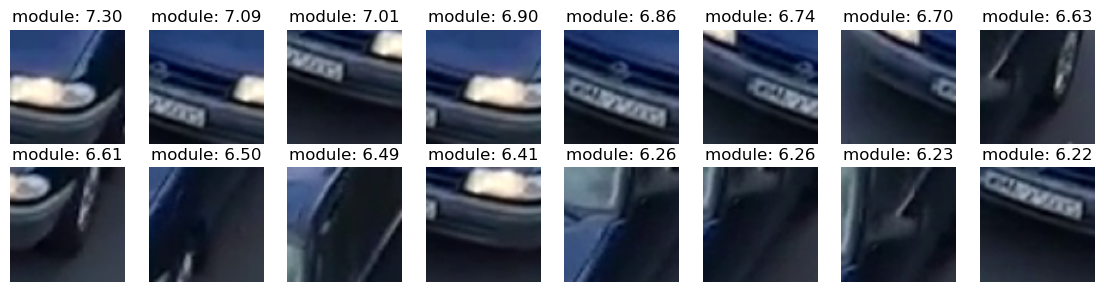}
    \caption{Sliding window size = 50 }
    \label{fig:short-d0}
  \end{subfigure}  
  \caption{Patches sorted by modules}
  \label{fig:metric}
\end{figure*}

With the subset \{car, bird, cat, deer, dog, horse, monkey\} of the STL-10 data set, where the car is the positive class, and the rest are negatives, carefully selecting three fixed $sws$ = 280, 180, 50, three iterations can be carried out in a one-car detection scenario \cref{fig:short-a0} according to Steps 1 to 4 of \cref{alg:algorithm 1}. Then by analyzing the positive patches sorted by module, as shown in \cref{fig:short-b0}, \cref{fig:short-c0}, and \cref{fig:short-d0}, we can find that it responds high modules to features such as plate, light, wheel when $sws$ = 50, features such as car head, body when $sws$ = 180, and the whole car feature, a combination of sub-features when $sws$ = 280. Therefore, the following proposition can be deduced.

\begin{proposition}
\label{pro:prob metric}
The module of the DisCNN's output vector is a metric of the presence probability of positive features. 
\end{proposition}

Based on \cref{pro:prob metric}, \cref{alg:algorithm 1} detects positive features, from concrete to abstract, by identifying the best-scoring patches across multi-scale sliding windows. Then, by overlapping these patches, a max-boundary bonding box can be drawn to localize the positive.

\subsection{One-car scenario: object detection via feature detection }
Using the same car-positive dataset as \cref{3.2}, Algorithm \ref{alg:algorithm 1} is fully executed this time. With a sliding window ranging from 400 to 40, the features of all scales are detected by a threshold of 6, as shown in \cref{fig:short-a}. By overlapping them, a maximum-boundary bonding box can be drawn to locate the positive, as shown in \cref{fig:short-b}. So \cref{alg:algorithm 1} is valid in object detection. 

\begin{figure*}
  \centering
  \begin{subfigure}{0.4\linewidth}
    \includegraphics[height=6cm]{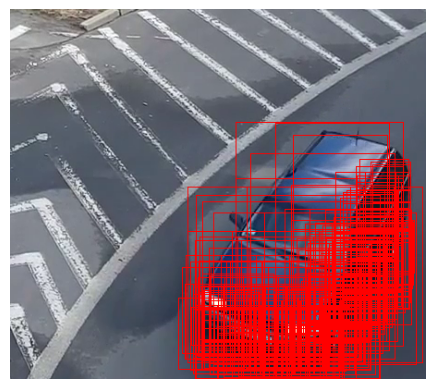}
    \caption{Bonding boxes of sliding windows from 400 to 40}
    \label{fig:short-a}
  \end{subfigure}
  \hspace{0mm}
  \begin{subfigure}{0.4\linewidth}
    \includegraphics[height=6cm]{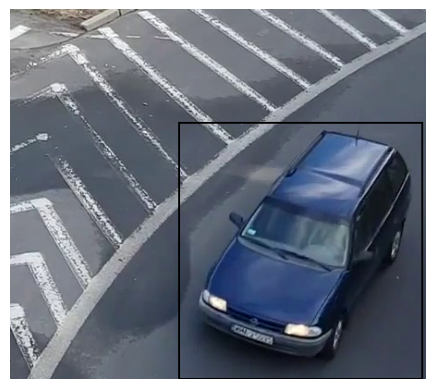}
    \caption{Bonding box with max boundary of \cref{fig:short-a}}
    \label{fig:short-b}
  \end{subfigure}\\
  \begin{subfigure}{0.4\linewidth}
    \includegraphics[height=6cm]{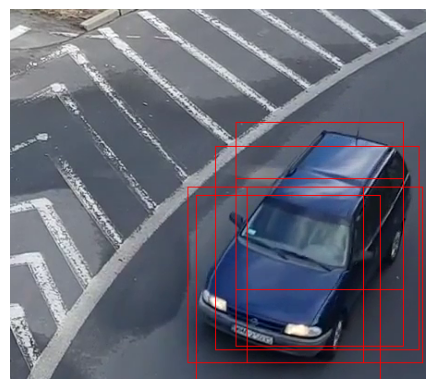}
    \caption{Bonding boxes of sliding windows from 220 to 180}
    \label{fig:short-c}
  \end{subfigure}
  \hspace{0mm}
  \begin{subfigure}{0.4\linewidth}
    \includegraphics[height=6cm]{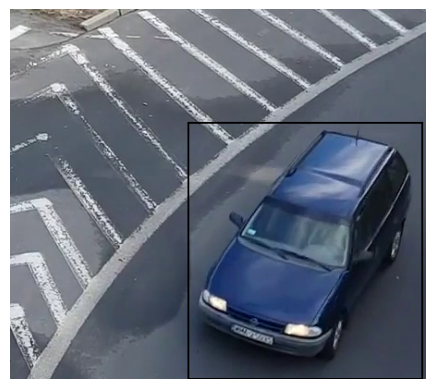}
    \caption{Bonding box with max boundary of \cref{fig:short-c}}
    \label{fig:short-d}
  \end{subfigure}
  \caption{One-car detection}
  \label{fig:fig 1}
\end{figure*}

\begin{proposition}
\label{pro:acc by feature detection}
DisCNN detects the positive by identifying its multi-scale features, from the positive itself to sub-features, thereby ensuring detection accuracy.
\end{proposition}

Moreover, if the car to be detected is of known size, the initial $sws$ and $min\_sws$ in Algorithm \cref{alg:algorithm 1} can be chosen appropriately to reduce the detection time. For example, changing the sliding window size range from [400, 40) to [220, 180) significantly reduces the detection time from 1.18 to 0.14 seconds. By a threshold of 5, the result is shown in \cref{fig:short-c} and \cref{fig:short-d}. 

\begin{figure*}
  \centering
  \begin{subfigure}{0.4\linewidth}
    \includegraphics[height=6cm]{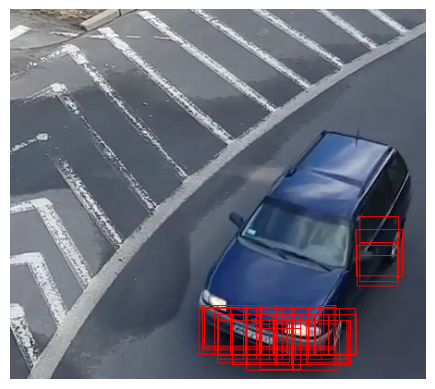}
    \caption{Bonding boxes of sliding windows from 60 to 40}
    \label{fig2:short-a}
  \end{subfigure}
  \hspace{0mm}
  \begin{subfigure}{0.4\linewidth}
    \includegraphics[height=6cm]{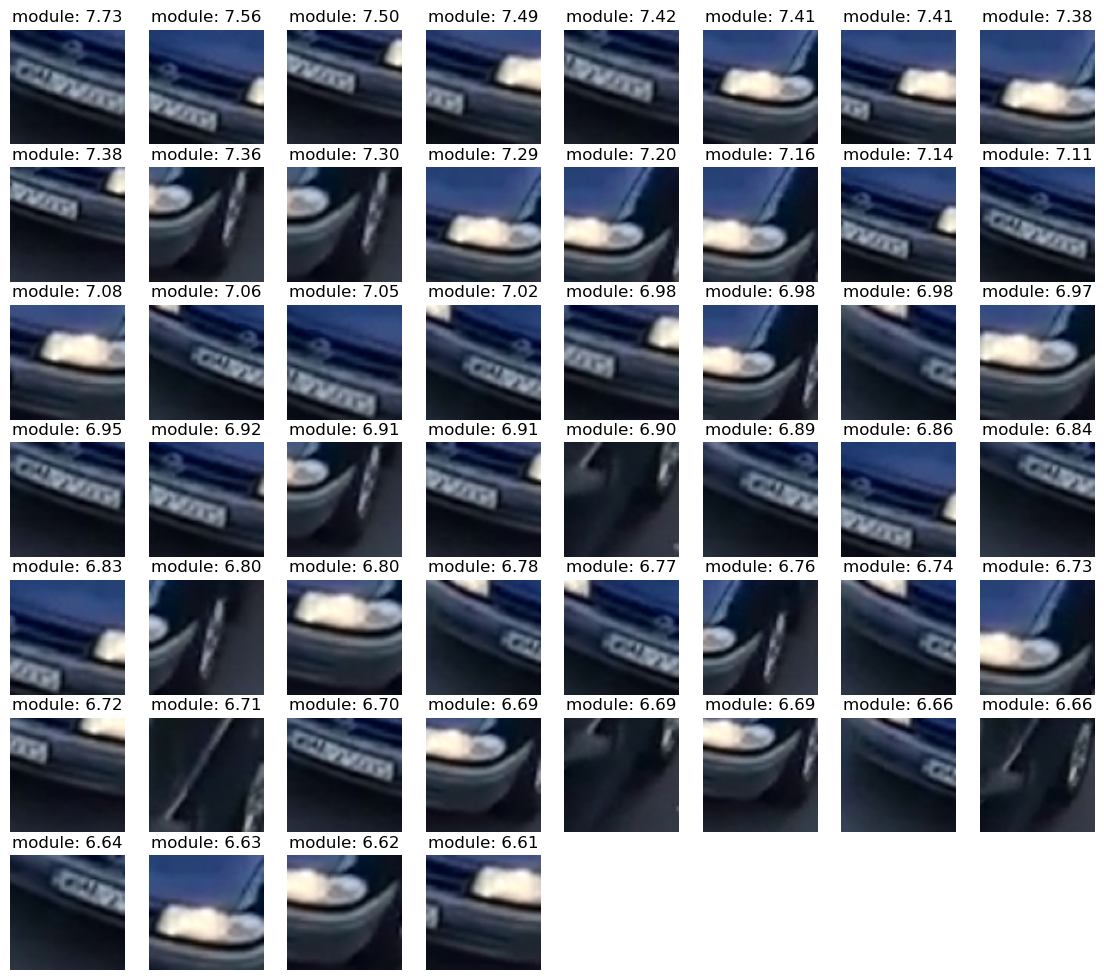}
    \caption{Car sub-features captured}
    \label{fig2:short-b}
  \end{subfigure}
  \caption{Car sub-feature detection}
  \label{fig2:fig 2}
\end{figure*}

Meanwhile, if the sliding window is adopted in the range [60, 40), which is relatively small compared to the car, with a threshold of 6.6, the result is shown in \cref{fig2:fig 2}. Therefore, it can detect car sub-features, such as license plates, lights, and wheels.

\begin{proposition}
\label{pro:feature detection}
DisCNN can be used for sub-feature detection of the positive class. 
\end{proposition}

\subsection{Multi-car scenario}

\begin{figure*}
  \centering
  \begin{subfigure}{0.66\linewidth}
    \includegraphics[height=6cm]{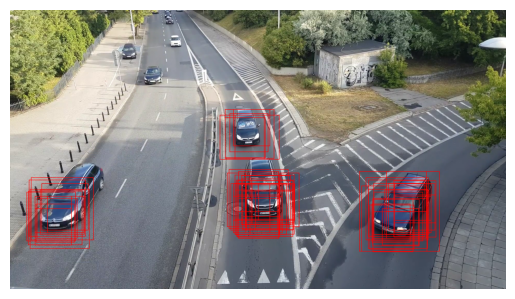}
    \caption{Bonding boxes of sliding windows from 400 to 140}
    \label{fig:short-a}
  \end{subfigure}\\
  \hspace{0mm}
  \begin{subfigure}{0.66\linewidth}
    \includegraphics[height=6cm]{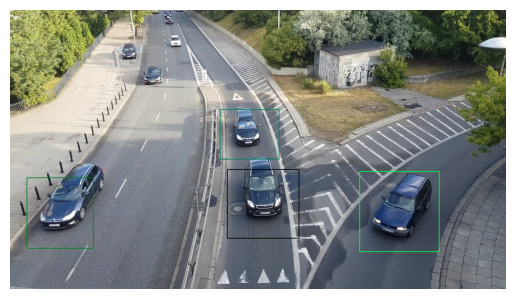}
    \caption{Bonding box with max boundary of \cref{fig:short-a}}
    \label{fig:short-b}
  \end{subfigure}\\
  \begin{subfigure}{0.4\linewidth}
    \includegraphics[height=6cm]{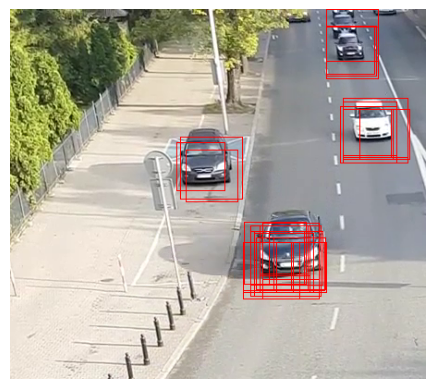}
    \caption{Bonding boxes of sliding windows from 140 to 50}
    \label{fig:short-c}
  \end{subfigure}
  \hspace{0mm}
  \begin{subfigure}{0.4\linewidth}
    \includegraphics[height=6cm]{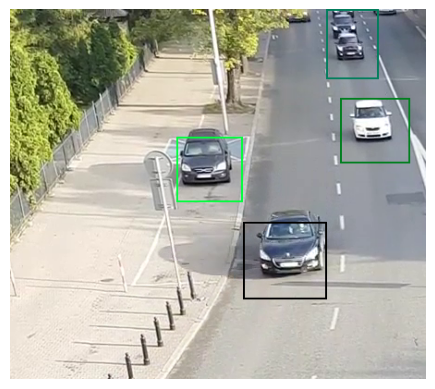}
    \caption{Bonding box with max boundary of \cref{fig:short-c}}
    \label{fig:short-d}
  \end{subfigure}
  \caption{multi-car detection}
  \label{fig:multi-car}
\end{figure*}

This time, the scenario of multiple cars is detected with \cref{alg:algorithm 1}. Using the same car-positive dataset as \cref{3.2}, with a sliding window ranging from 400 to 40, cars on different scales are detected by a threshold of 7, as shown in \cref{fig:multi-car}. Therefore, \cref{alg:algorithm 1} is still valid in multi-object detection. Note that the multi-car image is fairly large, and when the sliding windows are too small, the number of partitioned patches may be too large for our GPU memory. So when detecting relatively small cars with small sliding windows, we first have to cut out the subgraph containing them to avoid GPU memory failure. 

\section{Conclusion}
A multi-class object detection framework is presented based on an independent one-class object detection algorithm. All kinds of objects from different positive classes can be detected in parallel by DisCNNs trained separately on the corresponding datasets for each positive class. The most interesting fact is that with relatively small sliding windows, sub-features of the object can also be detected. This may indicate that the overall object is recognized by its components (i.e., sub-features), revealing the essential principle of DisCNN. For example, if only a light is captured by a small sliding window, the output module is fairly high, as shown in the sixth patch of \cref{fig2:short-b}. This module might stand for the distribution of cars, i.e., $p(car)$, in which $\textbf{P}(car) = \textbf{P}(car|light)\textbf{P}(light)$. In this case, $\textbf{P}(light)$ is close to 1, as the light almost occupies the entire patch, whose feature map is activated maximally. $\textbf{P}(car|light)$ is a prior probability distribution that is constant for all cars. In another case, the fourth patch of \cref{fig2:short-b} is considered, assume $\textbf{P}(car|light)$ and $\textbf{P}(car|light)$ that are known prior probabilities, and $\textbf{P}(car) = \textbf{P}(car|light)\textbf{P}(light)+\textbf{P}(car|plate)\textbf{P}(plate)$, where $\textbf{P}(light)$ and $\textbf{P}(plate)$ can be measured by how much the feature maps for them are activated. Therefore, we can assume that DisCNN recognizes objects by evaluating 
$\textbf{P}(object) = \sum_{i=1}^{n} \textbf{P}(object|feature_i)\textbf{P}(feature_i)$, the underlying principle of which would be an important research topic.

{
    \small
    \bibliographystyle{ieeenat_fullname}
    \bibliography{main}
}


\end{document}